\documentclass[letterpaper, 10 pt, conference]{ieeeconf}  

\IEEEoverridecommandlockouts                              

\usepackage{graphicx} 
\usepackage{times} 
\usepackage{amsmath} 
\usepackage{cite}
\usepackage[hidelinks]{hyperref}
\usepackage{booktabs}
\usepackage{caption}

\title{\LARGE \bf
Find Something You Can’t Do: Agentic Real-World Reinforcement Learning for Self-Improving VLA Models
}

\author{
Yuan Fang$^{1,*}$,
Zechu Li$^{1,2,*}$,
Haolei Tong$^{3}$,
Puze Liu$^{4,5}$,
Georgia Chalvatzaki$^{1,2}$
\thanks{
$^{*}$Equal contribution.
$^{1}$TU Darmstadt,
$^{2}$Hessian.AI,
$^{3}$University of Augsburg,
$^{4}$Tongji University,
$^{5}$Shanghai Research Institute for Intelligent Autonomous Systems.
This work was supported by Google.org, the DFG Excellence Cluster RAI, the Alfried Krupp Förderprize, the German Research Foundation (DFG) Emmy Noether Programme (CH 2676/1-1) and by ERC grant SIREN (101163933). Funded by the European Union. Views and opinions expressed are however those of the author(s) only and do not necessarily reflect those of the European Union or the European Research Council Executive Agency. Neither the European Union nor the granting authority can be held responsible for them.
The authors gratefully acknowledge the scientific support and HPC resources provided by the Erlangen National High Performance Computing Center (NHR\@FAU) of the Friedrich-Alexander-Universität Erlangen-Nürnberg (FAU) under the NHR project 80511. NHR funding is provided by federal and Bavarian state authorities.
}
}

\begin{document}

\maketitle
\thispagestyle{empty}
\pagestyle{empty}

\begin{abstract}
Vision--language--action (VLA) models provide strong priors for robotic manipulation but are typically deployed as frozen policies, unable to improve from their own failures. Real-world reinforcement learning (RL) offers a path to continued improvement, yet manual environment resets and task-success supervision hinder autonomous learning. We introduce \textbf{FIND}, an agentic real-world RL framework that closes the loop between scene understanding, weakness-aware practice, self-evaluation, and policy improvement in a persistent workspace. FIND reframes autonomous practice as a scene-conditioned, performance-aware task-selection problem: instead of restoring a predefined scene after each rollout, it uses the resulting scene to determine what to practice next. A vision--language agent identifies feasible tasks from a predefined library, prioritizes those with lower recent success rates, and evaluates outcomes using paired pre- and post-execution observations. We instantiate FIND with a frozen $\pi_{0.5}$ VLA and residual off-policy RL. Across eight real-world manipulation tasks, the independent human-assessed success rate improves from $55\%$ to $71.9\%$. A representative run completes 456 autonomous episodes within 6 hours of interaction, requiring 30 scene-recovery interventions and no human-provided reward labels during online learning. Ablations and systematic evaluations further examine key design choices, agent evaluation accuracy, and human intervention requirements. Our website is made publicly
available at: \href{https://fangyzzz.github.io/FIND.github.io/}{FIND.github.io}.

\end{abstract}
\section{Introduction}
Recent vision--language--action (VLA) models have made impressive progress toward general-purpose robotic manipulation, yet reliable real-world deployment remains an open challenge. Models such as RT-2, OpenVLA, $\pi_0$, and $\pi_{0.5}$~\cite{brohan2023rt,kim2024openvla,black2024pi_0,intelligence2025pi_} acquire semantic and visuomotor priors from large-scale robot datasets, but can still fail under changes in camera viewpoint, robot calibration, contact dynamics, workspace layout, or object geometry. A more fundamental limitation is that these models are typically deployed as frozen policies: when failures arise, they cannot learn from their own experience and instead require additional human-collected demonstrations and supervised fine-tuning. Real-world reinforcement learning (RL)~\cite{sutton1998reinforcement,kober2013reinforcement} offers a promising alternative by allowing a deployed VLA to collect experience, correct its failures, and improve directly through interaction.

\begin{figure*}[t]
    \centering
    \includegraphics[width=\textwidth]{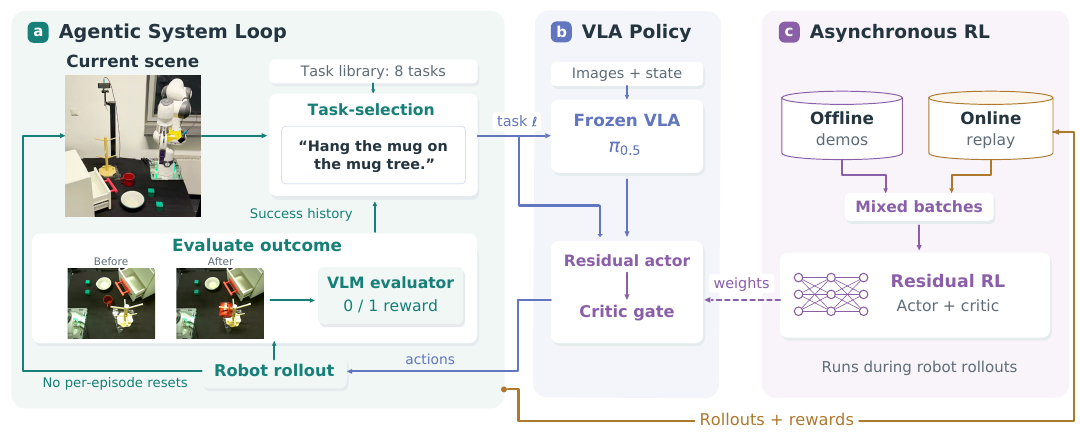}
    \vspace{-.4cm}
    \caption{
    Overview of \textbf{FIND}.
    (a) The agentic loop identifies feasible tasks in the persistent workspace, prioritizes practice based on recent task performance, and uses paired pre- and post-execution observations for VLM-based binary outcome evaluation.
    (b) A frozen VLA provides base actions, while a learned residual policy applies corrective actions through a critic-based execution gate.
    (c) Residual actor--critic learning runs asynchronously using mixed offline demonstrations and autonomously collected online experience, closing the loop between real-world practice, self-evaluation, and policy improvement.
    }
    \label{fig:overview}

\vspace{-.4cm}
\end{figure*}

Two major bottlenecks prevent real-world RL from operating autonomously: resetting the environment and evaluating task success. After each rollout, manipulated objects may be left in configurations that are unsuitable for the next attempt, requiring a human operator to restore the workspace. At the same time, determining whether an attempt succeeded often depends on task-specific reward instrumentation or manual labels. Prior work~\cite{eysenbach2017leave,gupta2021reset, sharma2023self} reduces these burdens through learned reset policies, forward--reverse behaviors, multi-task interaction, learned success classifiers, or vision--language-based evaluation. Recent advances in agentic systems, such as ASPIRE~\cite{lu2026aspire}, use large-language model (LLM) agents to interpret execution feedback and iteratively refine robot skills. Nevertheless, task selection, environment resetting, outcome evaluation, and policy improvement are commonly treated as separate processes, leaving sustained autonomous real-world learning an open challenge.

We argue that an autonomous robot need not reset the world if it can determine what to practice next. Indeed, a robot capable of resetting an arbitrary scene may already possess much of the general manipulation capability that it is supposed to acquire: \textit{restoring displaced objects from diverse failure states can be harder than completing the original task}. Rather than returning the workspace to a predefined configuration, an embodied agent can interpret the scene produced by its previous interaction, identify which tasks are now feasible, and select a task that exposes a current weakness of the policy. This reframes real-world learning from repeatedly resetting the environment into a scene-conditioned task-selection problem, where the outcome of one task becomes the starting condition for another.

Motivated by this, we introduce \textbf{FIND}, an agentic real-world RL framework that enables VLA self-improvement that runs an autonomous loop in a persistent workspace. Instead of restoring the scene after each episode, FIND treats the state left by one rollout as the starting point for the next. A vision--language agent identifies feasible tasks from a predefined library, with a
performance-aware curriculum prioritizing those with lower recent success rates, and uses paired-image evaluation to provide a binary outcome for both subsequent task allocation and policy learning. FIND is agnostic to the underlying policy-improvement algorithm; in this work, we instantiate its learning backend using a frozen $\pi_{0.5}$ VLA and residual off-policy RL trained from offline demonstrations and autonomously collected online experience. In an eight-task real-world workspace, FIND improves the independent human-assessed success rate from $55\%$ to $71.9\%$. In a representative 6-hour autonomous interaction run, the system requires 30 scene-recovery interventions and no human-provided reward labels during online learning.



The main contributions of this work are threefold:
\begin{itemize}
\item We introduce \textbf{FIND}, an agentic real-world RL framework that closes the autonomous loop between scene understanding, weakness-aware practice, self-evaluation, and policy improvement in a persistent workspace. 
\item We formulate the autonomous practice as a scene-conditioned, performance-aware task-selection problem, prioritizing tasks that are feasible in the current scene but not yet mastered, with self-evaluated outcomes updating both task selection and policy learning.
\item We show that FIND improves the independent human-assessed success rate from $55\%$ to $71.9\%$ across eight real-world tasks, with consistent trends in online VLM-based evaluation. A representative 6-hour autonomous interaction run further demonstrates low human intervention, together with ablations and analyses of VLM supervision and human intervention.
\end{itemize}

\section{Related Works}

\subsection{Real-World Reinforcement Learning for VLAs.}

Real-world reinforcement learning provides a mechanism for improving robot policies directly from deployment experience, but sample efficiency and stable adaptation remain key challenges~\cite{kober2013reinforcement}. Recent work has increasingly focused on adapting pretrained visuomotor and VLA policies through online reinforcement learning. RoboFuME~\cite{yang2024robot} combines offline-to-online RL with VLM-based reward classification for autonomous real-world fine-tuning, while residual RL~\cite{johannink2019residual,ankile2025residual} preserves a pretrained policy and learns corrective actions on top of its behavior. Xiao et al.~\cite{xiao2026self} extend residual RL to self-improving VLA policies through real-world interaction, and RLDG~\cite{xu2024rldg} uses task-specific RL policies to generate improved experience for training generalist robot policies. Complementary approaches explore other components of online VLA adaptation: RL Token~\cite{xu2026rl} reuses internal VLA representations for reinforcement learning, while Q2RL~\cite{dodeja2026life} uses learned Q-values to arbitrate between pretrained and RL actions during on-robot fine-tuning.

These approaches primarily address \emph{how} to improve a pretrained policy once the task and supervision are specified. \textbf{FIND} instead focuses on \emph{what to practice next}, coupling scene-conditioned feasibility, performance-aware task allocation, and autonomous outcome evaluation during persistent multi-task interaction.

\subsection{Autonomous and Reset-Free Robot Learning.}

Reset-free and autonomous robot learning aim to reduce the human intervention required for long-duration real-world RL. Leave No Trace~\cite{eysenbach2017leave} jointly learns task and reset policies, while Gupta et al.~\cite{gupta2021reset} exploit mutually compatible tasks so that the terminal state of one behavior can initialize another. VaPRL~\cite{sharma2021autonomous} constructs a curriculum over useful initial states, while Practice Makes Perfect~\cite{kumar2024practice} uses competence-aware planning to decide which parameterized skills should be practiced and autonomously improves them without environment resets. ReLMM~\cite{sun2022fully} enables autonomous navigation and manipulation without manual intervention. Demonstration-Bootstrapped Autonomous Practicing~\cite{gupta2022bootstrapped} uses prior demonstrations to bootstrap multi-task policies and task sequencing, while MEDAL++~\cite{sharma2023self} jointly learns task and undo behaviors. More recently, Robot-Trains-Robot~\cite{hu2025robot} uses a robotic teacher to provide automatic resets, rewards, safety support, and training schedules for real-world humanoid RL.

Prior reset-free approaches commonly rely on learned reset or undo behaviors, compatible task structures, or explicit curricula, while \textbf{FIND} formulates the autonomous and reset-free loop as an agentic system empowered by VLMs. 

\subsection{Agentic Systems for Robot Learning.}

Foundation models are increasingly used as high-level reasoning components for robot systems. SayCan~\cite{ichter2023saycan} grounds language-model plans with learned skill affordances, Inner Monologue~\cite{huang2022inner} incorporates environment feedback into closed-loop reasoning, and Code as Policies~\cite{liang2023code} generates executable robot programs. PaLM-E~\cite{driess2023palme} integrates language, vision, and embodied state, while Grounded Decoding~\cite{huang2023grounded} incorporates grounded affordance models into language-model generation. VoxPoser~\cite{huang2023voxposer} converts language instructions into 3D value maps for manipulation, Language to Rewards~\cite{yu2023language} generates reward functions from natural-language specifications, and KnowNo~\cite{ren2023robots} estimates planning uncertainty to determine when human assistance is required. Agentic Skill Discovery~\cite{zhao2025agentic} uses LLM-generated task proposals and reward functions together with RL and VLM verification to autonomously expand a robot's skill library. More recent systems such as HARBOR~\cite{li2026harbor} and Nautilus~\cite{jin2026nautilus} extend foundation-model reasoning toward broader agentic robot-learning workflows. In contrast, \textbf{FIND} focuses on the agentic system for real-world RL to enable self-improving VLA models, differing from prior works in the targeted problem.

\section{Methods}

We propose \textbf{FIND}, an agentic framework for autonomous VLA self-improvement through persistent real-world interaction. Rather than repeatedly restoring the workspace after each rollout, FIND treats the resulting scene as the starting point for subsequent practice. A VLM identifies tasks that are feasible in the current scene, while a performance-aware curriculum prioritizes feasible tasks according to the recent competence of the policy. After execution, paired-image self-evaluation determines task success and uses the resulting outcome both to update future task allocation and to provide a learning signal for policy improvement. In our implementation, policy improvement is instantiated using a frozen VLA together with residual off-policy reinforcement learning over offline demonstrations and autonomously collected experience. An overview of FIND is shown in Fig.~\ref{fig:overview}.

\subsection{Problem Formulation}
\label{sec:problem_formulation}

We formulate autonomous self-improvement as repeatedly selecting and learning from tasks in a persistent multi-task workspace. We assume that (i) the workspace admits multiple language-conditioned tasks whose feasibility varies with the scene state, and (ii) the pretrained VLA has non-trivial competence on these tasks, such that online RL refines existing behaviors rather than discovering them from scratch. Let

\vspace{-.7cm}
\begin{equation}
\mathcal{T}
=
\{\tau_1,\ldots,\tau_N\}
\end{equation}
denote a predefined library of language-conditioned manipulation tasks, where each task $\tau_i$ is associated with a natural-language instruction $l_i$. At control step $t$, the robot receives an observation\vspace{-0.3cm}
\begin{equation}
o_t
=
\left(
\mathcal{I}_t,
s_t,
l_i
\right),
\end{equation}
where $\mathcal{I}_t$ denotes the visual observations and $s_t$ denotes the robot proprioceptive state. A language-conditioned policy $\pi$ maps the observation to a robot action,
\begin{equation}
a_t \sim \pi(\cdot\mid o_t).
\end{equation}

Persistent interaction changes the conventional episodic learning problem because the workspace is not manually restored after each rollout. Let $z_k$ denote the task-relevant workspace state at the beginning of episode $k$. Since different scene configurations admit different manipulation opportunities, only a subset of the task library may be feasible,
\vspace{-0.2cm}

\begin{equation}
\mathcal{F}(z_k)
=
\left\{
\tau_i\in\mathcal{T}
\mid
\tau_i \text{ is feasible under } z_k
\right\}.
\end{equation}
\vspace{-0.4cm}

FIND selects a task $\tau_k\in\mathcal{F}(z_k)$, executes the corresponding language-conditioned policy, and obtains a binary task outcome
\vspace{-0.2cm}
\begin{equation}
r_k\in\{0,1\},
\end{equation}
\vspace{-0.5cm}

indicating whether the task was successfully completed. Rather than manually restoring the manipulated objects, the physical configuration produced by the rollout defines the workspace state $z_{k+1}$ for the subsequent episode. The interaction therefore follows

\vspace{-0.3cm}
\begin{equation}
z_k
\rightarrow
\mathcal{F}(z_k)
\rightarrow
\tau_k
\rightarrow
r_k
\rightarrow
z_{k+1}.
\end{equation}
\vspace{-0.4cm}

This formulation exposes three decisions required for autonomous self-improvement: determining \emph{what can be practiced} in the current scene, deciding \emph{what should be practiced} given the current competence of the policy, and determining \emph{whether the resulting attempt succeeds}. FIND addresses these decisions through the agentic system described next.

\begin{figure*}[t]
  \centering
  \includegraphics[width=\textwidth]{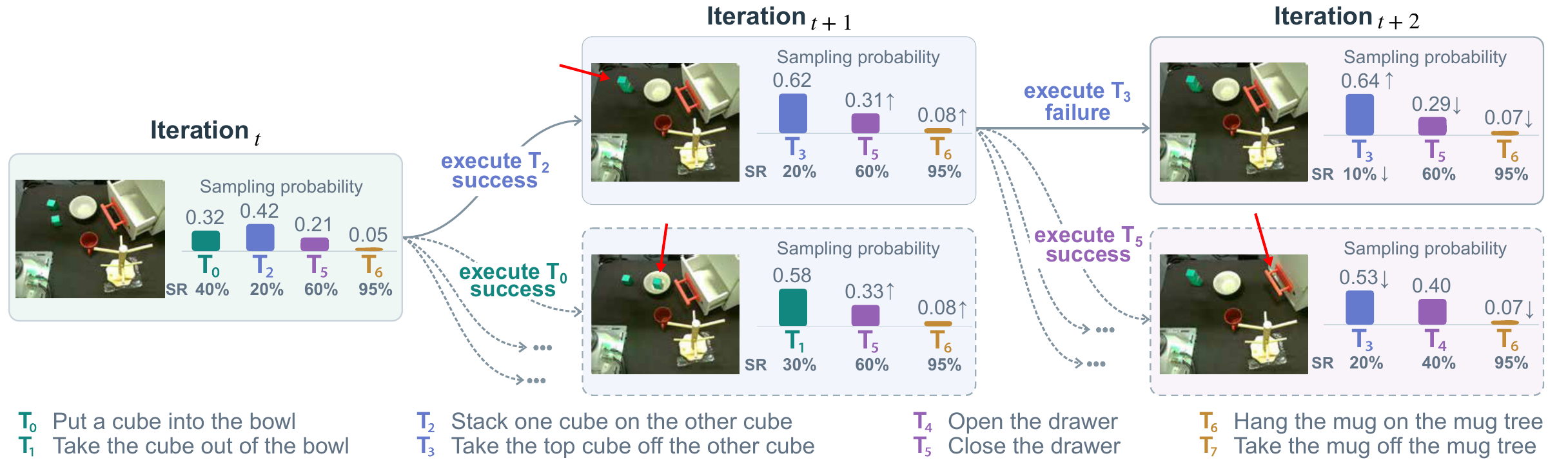}
  \vspace{-0.4cm}
  \caption{
  Illustration of scene-conditioned, performance-aware task sampling in FIND.
  At each iteration, the current workspace determines the feasible task set, while recent task success rates determine the sampling probabilities among feasible tasks.
  After executing a sampled task, the resulting scene and self-evaluated outcome update task feasibility and performance history, producing a new sampling distribution for the next iteration.
  }
  \label{fig:task_sampling}
\vspace{-0.4cm}
\end{figure*}

\subsection{Agentic Self-Improvement System}
\label{sec:agentic_system}

FIND closes the autonomous interaction loop through scene-conditioned task selection, performance-aware task allocation, and paired-image self-evaluation, as illustrated in Fig.~\ref{fig:overview}(a). Together, these components allow the robot to use the state produced by one rollout directly as the starting condition for subsequent practice, while continuously adapting what it practices according to its recent performance.

\paragraph{Scene-conditioned task selection}
Scene-conditioned task selection allows FIND to continue practicing without repeatedly restoring the manipulated objects to a predefined configuration. At the beginning of episode $k$, the VLM analyzes the current workspace observation and infers the task-relevant scene state $z_k$, as illustrated in Fig.~\ref{fig:task_sampling}. The feasible subset $\mathcal{F}(z_k)$ is then determined from the predefined task library, and tasks outside this subset are excluded from the next interaction.

The task library and feasibility relations are predefined rather than freely generated by the VLM agent. This keeps the selected instructions within the task space supported by the pretrained VLA and makes the possible transitions between workspace configurations explicit. Importantly, the robot does not need to reverse the previous action or restore a particular initial state: any task that is valid under the resulting scene can become the next practice task.

\paragraph{Performance-aware curriculum}
The performance-aware curriculum determines which feasible task should receive the next unit of real-world interaction. A fixed or uniform distribution does not account for the evolving competence of the policy and may continue allocating expensive interaction to tasks that are already solved reliably. FIND instead tracks recent task performance and gives higher priority to tasks on which the current policy performs poorly.

For every task $\tau_i$, FIND maintains a rolling history of binary task outcomes,\vspace{-0.2cm}
\begin{equation}
\mathcal{H}_i
=
\left[
r_i^{(1)},
r_i^{(2)},
\ldots,
r_i^{(K)}
\right],
\qquad
r_i^{(j)}\in\{0,1\},
\end{equation}
\vspace{-0.5cm}

where $K$ denotes the curriculum window size. We set \(K=20\) for all experiments. The recent success rate is estimated as\vspace{-0.2cm}
\begin{equation}
\hat{p}_i
=
\frac{1}{|\mathcal{H}_i|}
\sum_{r\in\mathcal{H}_i} r.
\end{equation}
\vspace{-0.4cm}

The curriculum is initialized from the competence of the pretrained base policy rather than from an uninformed uniform prior. Before online self-improvement begins, each task history is populated using base-policy evaluation rollouts. During autonomous training, every newly self-evaluated outcome is appended to the corresponding history and the oldest entry is removed. The initialization outcomes are therefore gradually replaced by online experience, allowing $\hat p_i$ to track the recent competence of the evolving policy.

Task priority is derived directly from the recent success estimate. We assign\vspace{-0.3cm}
\begin{equation}
w_i
=
\max
\left(
\epsilon,
1-\hat{p}_i
\right),
\end{equation}
\vspace{-0.6cm}

where $\epsilon>0$ defines a minimum sampling weight. Tasks with lower recent success rates receive higher weights, while the lower bound prevents reliably solved tasks from being permanently excluded and allows the system to continue detecting performance regressions.

Scene feasibility and policy competence jointly determine the final task-sampling distribution. Since a high-priority task may not be physically executable in the current workspace, FIND masks all infeasible tasks and normalizes the remaining weights,\vspace{-0.2cm}
\begin{equation}
P(\tau_i\mid z_k)
=
\begin{cases}
\dfrac{w_i}
{\sum_{\tau_j\in\mathcal{F}(z_k)}w_j},
&
\tau_i\in\mathcal{F}(z_k),\\[8pt]
0,
&
\tau_i\notin\mathcal{F}(z_k).
\end{cases}
\end{equation}
\vspace{-0.4cm}

The resulting distribution changes continuously as both the workspace and the policy evolve. As illustrated in Fig.~\ref{fig:task_sampling}, a rollout changes the physical scene and therefore the set of feasible tasks, while its outcome changes the corresponding performance estimate and sampling weight. FIND consequently prioritizes \emph{what the robot still needs to practice} among \emph{what it can currently execute}, rather than following a fixed multi-task schedule.

\paragraph{Paired-image self-evaluation}
Paired-image self-evaluation provides autonomous task-level supervision without requiring human outcome labels. Immediately before executing the selected task, FIND records an observation of the workspace. After the rollout terminates, a second observation is captured from the same scene-level viewpoint. The VLM agent receives the task instruction together with the pre- and post-execution observations and predicts a binary outcome,
\begin{equation}
r_k
=
\begin{cases}
1,
& \text{if the task is successfully completed},\\
0,
& \text{otherwise}.
\end{cases}
\end{equation}
\vspace{-0.35cm}

Using paired observations allows the evaluator to reason about the state transition produced by the current rollout rather than relying only on the final scene. This is particularly useful when the success criterion depends on how an object or articulated component changed relative to its initial configuration.

The self-evaluated outcome connects autonomous supervision to both current learning and future practice. First, $r_k$ provides a sparse reward used by the policy-improvement backend. Second, it is inserted into the corresponding history $\mathcal{H}_i$, updating the performance-aware curriculum for subsequent task selection. The same outcome therefore closes the loop between task execution, self-evaluation, data allocation, and policy improvement.

\subsection{Policy Improvement via Residual Reinforcement Learning}
\label{sec:residual_learning}

The FIND interaction loop is independent of the specific policy-improvement algorithm. In this work, we instantiate the learning backend using residual off-policy reinforcement learning~\cite{ankile2025residual,xiao2026self}, as illustrated in Fig.~\ref{fig:overview}(b--c). The pretrained VLA remains frozen, while a lightweight residual policy learns bounded corrections from offline demonstrations and autonomously collected experience.

\paragraph{Residual policy}
Given visual observations $\mathcal{I}_t$, proprioceptive state $s_t$, and instruction $l$, the frozen VLA produces the base action
\vspace{-0.2cm}
\begin{equation}
\mathbf{a}^{\mathrm{base}}_t
=
\pi_{\mathrm{VLA}}(\mathcal{I}_t,s_t,l).
\end{equation}
\vspace{-0.5cm}

The residual actor predicts a correction from visual features and a language-conditioned robot state,
\vspace{-0.2cm}
\begin{equation}
\Delta\mathbf{a}_t
=
\pi_{\phi}
\left(
h_{\mathrm{img}}(\mathcal{I}_t),
\tilde{s}_t
\right),
\end{equation}
\vspace{-0.5cm}

and the combined action is
\vspace{-0.2cm}
\begin{equation}
\mathbf{a}^{\mathrm{comb}}_t
=
\operatorname{clip}
\left(
\mathbf{a}^{\mathrm{base}}_t
+
\alpha\Delta\mathbf{a}_t
\right),
\end{equation}
\vspace{-0.5cm}

where $\alpha$ bounds the residual magnitude. The residual actor is initialized to output zero correction, initially reproducing the frozen VLA behavior.

Following prior residual off-policy RL~\cite{ankile2025residual,xiao2026self}, we train with mixed batches of offline demonstrations and autonomously collected trajectories. Offline base actions are reconstructed by querying the frozen VLA. We use a REDQ-style ensemble critic~\cite{chen2021randomized} with ten Q-value heads to improve value-estimation robustness, with the ensemble mean used for actor optimization and execution gating.

As illustrated in Fig.~\ref{fig:overview}(c), we improve real-world training efficiency by executing rollout collection and RL optimization asynchronously in separate Python threads, allowing robot interaction and policy learning to proceed concurrently. The collector transfers transitions to the learner through an ordered FIFO queue, while updated parameters are returned through a latest-only queue so that the collector uses the most recent available policy. To bound learner lag, if the collector becomes more than 2000 environment steps ahead, it completes the current episode and pauses until the learner catches up. This design reduces idle time between interaction and optimization while preserving ordered experience transfer and bounded policy staleness.

\paragraph{Language conditioning}
The residual learner uses an independent language branch rather than internal VLA representations. A frozen Sentence-BERT encoder~\cite{reimers2019sentence} followed by a trainable projection produces
\vspace{-0.25cm}
\begin{equation}
z_l
=
g_{\theta_l}\!\left(f_{\mathrm{SBERT}}(l)\right),
\qquad
\tilde{s}_t=[s_t;z_l].
\end{equation}
\vspace{-0.5cm}

The resulting language-conditioned state is provided with visual features to both actor and critic. The projection is optimized through the critic objective and detached during actor updates.

\begin{table}[t]
\centering
\begin{tabular}{l|l|l} 
\toprule
\textbf{Task}                        & \textbf{Episodes} & \textbf{Transitions}  \\  
\midrule
Put a cube into the bowl                     & 50                  & 6,573            \\ 
\midrule
Take the cube out of the bowl        & 50                   & 5,922             \\ 
\midrule
Stack one cube on the other cube     & 50                  & 8,228            \\ 
\midrule
Take the top cube off the other cube & 50                   & 6,364             \\ 
\midrule
Open the drawer                      & 50                   & 6,534              \\ 
\midrule
Close the drawer                     & 50                   & 3,974              \\ 
\midrule
Hang the mug on the mug tree         & 50                  & 8,930             \\ 
\midrule
Take the mug off the mug tree        & 50                   & 16,384            \\
\bottomrule

\end{tabular}
\vspace{0.05cm}
\caption{Offline dataset composition across the eight reversible manipulation tasks.}
\label{tab:dataset}
\vspace{-0.9cm}

\end{table}

\paragraph{Critic-gated residual execution}
To prevent unreliable residual corrections from degrading the pretrained behavior, FIND compares the predicted values of the base and corrected actions. Let
\vspace{-0.2cm}
\begin{equation}
x_t
=
\left(
h_{\mathrm{img}}(\mathcal{I}_t),
\tilde{s}_t
\right)
\end{equation}
\vspace{-0.5cm}

denote the critic state representation. The estimated residual advantage is
\vspace{-0.15cm}
\begin{equation}
A_{\mathrm{res}}(x_t)
=
\bar{Q}
\left(
x_t,
\mathbf{a}^{\mathrm{comb}}_t
\right)
-
\bar{Q}
\left(
x_t,
\mathbf{a}^{\mathrm{base}}_t
\right),
\end{equation}
\vspace{-0.5cm}

where $\bar{Q}$ denotes the ensemble-averaged critic value. The executed action is
\vspace{-0.2cm}
\begin{equation}
\mathbf{a}^{\mathrm{exec}}_t
=
\begin{cases}
\mathbf{a}^{\mathrm{comb}}_t,
&
A_{\mathrm{res}}(x_t)>\delta,\\[3pt]
\mathbf{a}^{\mathrm{base}}_t,
&
\text{otherwise},
\end{cases}
\end{equation}

where $\delta$ is the gating threshold. The gate falls back to the pretrained VLA whenever the critic does not predict sufficient benefit from the residual correction.

\section{Experiments}
\begin{figure*}[t]
    \centering
    \includegraphics[width=\textwidth]{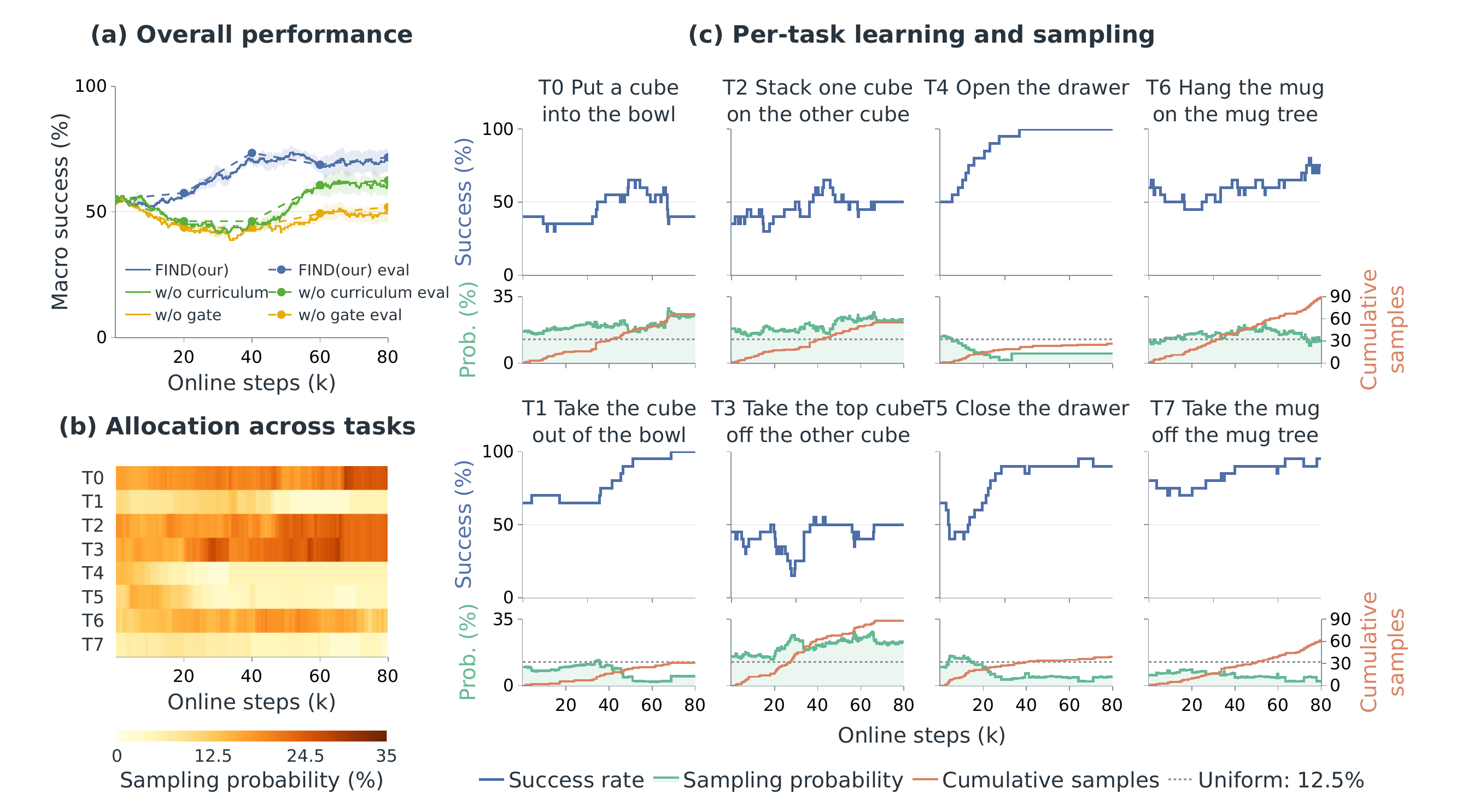}
    \vspace{-0.7cm}

    \caption{
Autonomous multi-task self-improvement with FIND.
(a) Online rolling success rates (solid) are reported as the mean across five independent runs, with shaded regions indicating $\pm1$ standard deviation across runs. Independent human-assessed evaluation results, conducted every 20k online steps, are shown as dashed lines.
(b) Evolution of the task-sampling distribution across the eight manipulation tasks on one representative run.
(c) Per-task success rate, sampling probability, and cumulative samples for one representative run.
The curriculum reallocates interaction toward tasks with lower recent success while reducing sampling of reliable tasks.
}
    \label{fig:overview results}
    \vspace{-0.5cm}

\end{figure*}
Our experiments investigate whether FIND can improve a pretrained VLA through autonomous real-world interaction with limited human intervention, and how performance-aware task selection and critic gating contribute to this improvement. Section~\ref{exp:A} introduces the robotic platform, eight manipulation tasks, and the demonstration dataset used for base-policy fine-tuning and replay initialization. Section~\ref{exp:B} evaluates autonomous operation and policy improvement, analyzes how task allocation evolves during learning, and presents ablation studies to assess the contributions of the curriculum and critic gate.
\subsection{Experimental Setup and Dataset}
\label{exp:A}

All experiments are conducted in a real-world tabletop environment using a 7-DoF Franka Emika Panda equipped with a parallel-jaw gripper. The workspace contains two cubes, a bowl, a drawer, a mug, and a mug tree, forming eight manipulation tasks organized into four reversible pairs: cube insertion/removal, stacking/unstacking, drawer opening/closing, and mug hanging/removal. This structure allows the resulting scene of one rollout to provide feasible starting conditions for subsequent tasks.

The perception system uses two ZED X Mini cameras, with one third-person and one wrist-mounted view. Only the left RGB streams are used and resized to $224\times224$. We fine-tune $\pi_{0.5}$~\cite{intelligence2025pi_} as the base VLA using real-world demonstrations. The policy receives the two RGB observations, language instruction, and robot proprioception, and outputs an 8-D end-effector pose and gripper command. During online learning, the frozen VLA action is combined with the bounded correction predicted by the residual policy.

Demonstrations are collected using a Meta Quest controller and stored in the LeRobot format~\cite{cadene2024lerobot}. Each trajectory contains RGB observations, robot proprioception, end-effector and gripper actions, language instructions, and terminal information. Table~\ref{tab:dataset} summarizes the offline dataset, which is used both to fine-tune the base VLA and to initialize the offline replay buffer. Experiments run on a workstation with an NVIDIA RTX 4090 GPU, an Intel Core i9-14900K CPU, and 64 GB RAM.

\subsection{Experimental Results}
\label{exp:B}
Fig.~\ref{fig:overview results} summarizes the autonomous learning behavior of FIND. We evaluate both the overall improvement of the multi-task policy and how task-sampling probabilities evolve in response to recent task success rates.

\begin{table}[t]
\centering
\small
\begin{tabular*}{\columnwidth}{@{\extracolsep{\fill}}lr@{}}
\toprule
\textbf{Operation statistics} & \textbf{Value} \\
\midrule
Total interaction time                 & 6 h    \\
Total episodes                         & 456      \\
Episodes without human intervention    & 426      \\
\textbf{Manual scene-recovery interventions}
                                       & \textbf{30} \\
\quad Object dropped in unfavorable pose & 21     \\
\quad Difficult-to-separate objects      & 7     \\
\quad Object outside reachable workspace & 2     \\
\midrule
Human-intervention episode ratio       & 6.58\%  \\
Longest uninterrupted autonomous run   & 48 min   \\
\bottomrule
\end{tabular*}
\vspace{0.1cm}
\caption{Autonomous operation statistics of FIND from a representative
6-hour real-world self-improvement run.
Human intervention denotes manual scene recovery required
for interaction to continue. The three indented categories
partition the 30 interventions.}
\label{tab:auto_eval}
\vspace{-0.9cm}
\end{table}

\paragraph{Autonomous operation}
Table~\ref{tab:auto_eval} reports autonomy statistics from a representative 6-hour self-improvement run comprising 456 real-world episodes. All manual scene-recovery events and their failure categories are verified by human evaluation, with an intervention counted only when manual scene recovery is required before autonomous interaction can continue. Of the 456 episodes, 426 proceeded without human intervention, while 30 required manual scene recovery, corresponding to an intervention rate of 6.58\%. The longest uninterrupted autonomous run lasted 48 minutes. Among the 30 interventions, 21 involved an object dropped into an unfavorable pose, 7 involved objects entering difficult-to-separate configurations, and 2 resulted from an object leaving the robot's reachable workspace. These cases mainly occurred when failures moved objects into configurations poorly covered by the training data, such as unusual mug poses or tightly coupled cube arrangements, where the underlying VLA had limited recovery competence. This highlights a remaining dependence of FIND on the recovery-state coverage of the pretrained policy.


\paragraph{VLM agent evaluation}
We compare five advanced multimodal models for the two agentic components of FIND: GPT-5.6 Terra~\cite{openai2026gpt56terra}, GPT-5.6 Luna~\cite{openai2026gpt56luna}, Claude Fable 5~\cite{anthropic2026fable5}, Gemini 3.1 Pro Preview~\cite{google2026gemini31pro}, and Kimi K3~\cite{moonshot2026kimik3}. Each model is evaluated on 100 scene observations for feasible-task selection and 100 pre/post-execution image pairs for paired-image success evaluation against human-annotated labels. Table~\ref{tab:vlm_eval} reports both accuracy and average response latency for the two components, with latency reported as mean $\pm$ standard deviation over 100 queries. GPT-5.6 Terra achieves the strongest overall accuracy, attaining 96\% on feasible-task selection and 100\% on paired-image success evaluation, while also exhibiting the lowest average latency among the evaluated models. Considering both evaluation accuracy and response efficiency, we use GPT-5.6 Terra for both supervisory components in the subsequent real-world experiments.

\paragraph{Overall self-improvement}
As shown in Fig.~\ref{fig:overview results}(a), FIND progressively improves performance across the eight manipulation tasks through autonomous interaction. For each method, we conduct five independent runs with different random seeds. For the online rolling success rate, we compute the mean and standard deviation across runs at each recorded training step. The solid curves report the mean, with shaded regions indicating $\pm 1$ standard deviation across runs. Independent human-assessed evaluations are conducted every 20k online steps and are shown as dashed lines. For each run and task, independent evaluation uses 20 human-assessed rollouts at each checkpoint, with success rates averaged across the eight tasks and then across the five runs. The online rolling success rate is computed over the most recent $K=20$ VLM-predicted binary outcomes. These evaluation rollouts are excluded from the replay buffer and curriculum statistics. Starting from 55\% for the fine-tuned base VLA, the mean independent human-assessed success rate across the five runs reaches 71.9\% at the final checkpoint. The online VLM-evaluated rolling success rate shows a consistent improvement trend, reaching $(70.4 \pm 4.6)\%$ at the final checkpoint. Removing either the performance-aware curriculum or the critic-based gate leads to substantially weaker improvement, indicating that effective self-improvement depends on both informative task allocation and conservative residual execution.

\begin{table}[t]
\centering
\resizebox{\columnwidth}{!}{%
\begin{tabular}{lcccc}
\toprule
& \multicolumn{2}{c}{\textbf{Task Selection}}
& \multicolumn{2}{c}{\textbf{Success Evaluation}} \\
\cmidrule(lr){2-3}\cmidrule(lr){4-5}
\textbf{Model}
& \textbf{Acc. (\%)}
& \textbf{Lat. (s)}
& \textbf{Acc. (\%)}
& \textbf{Lat. (s)} \\
\midrule
GPT-5.6 Terra
& \textbf{96}
& $\mathbf{4.1 \pm 1.2}$
& \textbf{100}
& $\mathbf{2.6 \pm 1.2}$ \\

GPT-5.6 Luna
& 88
& $4.4 \pm 1.7$
& 98
& $2.7 \pm 1.3$ \\

Claude Fable 5
& \textbf{96}
& $7.3 \pm 1.5$
& 97
& $5.8 \pm 0.9$ \\

Gemini 3.1 Pro Preview
& 94
& $9.2 \pm 3.1$
& 95
& $6.4 \pm 0.8$ \\

Kimi K3
& 93
& $33.5 \pm 22.1$
& 96
& $16.9 \pm 6.4$ \\
\bottomrule
\end{tabular}
}
\vspace{0.2cm}
\caption{VLM supervisory-component evaluation over 100 task-selection scenes and 100 pre/post-execution image pairs. We report accuracy and average response latency, with the best values in each column in bold.}

\label{tab:vlm_eval}
\vspace{-0.8cm}
\end{table}

\paragraph{Performance-aware task allocation}
Fig.~\ref{fig:overview results}(b--c) illustrates how FIND dynamically reallocates interaction according to scene feasibility and recent task performance. Rather than sampling uniformly across the eight tasks, the curriculum assigns higher probabilities to feasible tasks with lower recent success rates, while reducing the priority of tasks that become consistently successful. As reflected in the per-task learning curves, tasks that become more reliable are sampled less frequently, whereas tasks with lower or fluctuating success rates continue to receive additional interaction. The resulting distribution therefore evolves throughout training rather than remaining fixed, allowing practice to shift toward newly emerging weaknesses as the policy improves. This closed-loop relationship between self-evaluation, task allocation, and policy improvement is central to FIND: the same task outcomes used for reinforcement-learning supervision also determine where future real-world interaction is allocated.

\paragraph{Effect of curriculum and critic gating}
The ablations in Fig.~\ref{fig:overview results}(a) highlight the complementary roles of the two stabilization mechanisms. Without the performance-aware curriculum, interaction is not systematically directed toward the current weaknesses of the policy, resulting in markedly slower improvement. Removing the critic gate similarly reduces performance, as residual corrections are applied even when the critic does not predict an advantage over the frozen VLA action. The full model achieves the strongest and most consistent improvement, indicating that adaptive data allocation and conservative residual execution address different failure modes of autonomous real-world learning.
\section{Conclusion}

We presented \textbf{FIND}, an agentic real-world RL framework that enables a pretrained VLA to autonomously identify feasible tasks, prioritize its current weaknesses, evaluate task outcomes, and improve through continued interaction. By treating the resulting workspace state as the starting point for the next episode, FIND replaces repeated manual object resetting with scene-conditioned task selection. A performance-aware curriculum and paired-image self-evaluator close the loop between autonomous data collection, task selection, and policy improvement. Across eight reversible real-world manipulation tasks, FIND continuously reallocates interaction toward feasible tasks with lower recent success rates and improves the deployed policy while substantially reducing manual resets and human-provided task labels. These results suggest that reasoning about \emph{what to practice next} provides a practical path toward more autonomous real-world robot learning.

\textbf{Limitations.}
The current formulation operates within a predefined task space that is largely supported by the pretrained VLA, and therefore assumes that the base policy has sufficient prior competence to attempt the available tasks. In addition, our task suite is designed around reversible or mutually compatible task pairs so that the outcome of one interaction can provide a valid starting configuration for another. While this structure substantially reduces manual object resetting, it may not directly extend to tasks whose terminal states do not naturally enable subsequent useful interactions. More generally, autonomous practice is constrained by how well the available task library covers the workspace configurations that may arise during deployment.

\textbf{Future Work.}
Future work can relax these assumptions by extending FIND to larger and more open-ended task spaces, where the robot can autonomously discover new practice opportunities rather than selecting only from a fixed library. An important direction is to reduce the reliance on explicitly paired tasks by allowing the agent to reason over longer sequences of skills, discover intermediate recovery or rearrangement behaviors, and construct its own transitions between useful workspace states. Another direction is to jointly expand the capabilities of the underlying VLA during deployment, so that the set of tasks available for autonomous practice can grow as new behaviors are acquired. Ultimately, this could move FIND from identifying weaknesses within an existing repertoire toward continuously expanding what the robot is able to practice and learn in the real world.

\bibliographystyle{IEEEtran}
\bibliography{main}

\end{document}